\documentclass[nonacm,sigconf]{acmart}
\AtBeginDocument{%
  }

\copyrightyear{2025}
\acmYear{2025}
\acmConference[Arxiv]{}{Nov}{2025}
\acmISBN{}

\usepackage{amsthm}
\newtheorem{definition}{Definition}

\usepackage{graphicx}  
\usepackage{makecell}  
\usepackage{wrapfig}
\usepackage{subcaption}
\usepackage{siunitx} 
 \usepackage{multirow}

\usepackage{amssymb} 
\usepackage{pifont}       
\usepackage{xcolor}       
\usepackage{rotating}   
\usepackage{booktabs}   
\usepackage{adjustbox}  
\usepackage{algorithm}
\usepackage{algpseudocode}

\begin{document}

\title[Physically Interpretable World Models via Weakly Supervised Representation Learning]{Physically Interpretable World Models via \\Weakly Supervised Representation Learning}



\author{Zhenjiang Mao}
\affiliation{%
  \institution{University of Florida}
  \city{Gainesville, FL}
  \country{USA}
}

\author{Mrinall Eashaan Umasudhan}
\affiliation{%
  \institution{University of Florida}
  \city{Gainesville, FL}
  \country{USA}
}

\author{Ivan Ruchkin}
\affiliation{%
  \institution{University of Florida}
  \city{Gainesville, FL}
  \country{USA}
}




\renewcommand{\shortauthors}{Mao et al.}

\begin{abstract}
Learning predictive models from high-dimensional sensory observations is fundamental for cyber-physical systems, yet the latent representations learned by standard world models lack physical interpretability. This limits their reliability, generalizability, and applicability to safety-critical tasks. We introduce Physically Interpretable World Models (PIWM), a framework that aligns latent representations with real-world physical quantities and constrains their evolution through partially known physical dynamics. Physical interpretability in PIWM is defined by two complementary properties: (i) the learned latent state corresponds to meaningful physical variables, and (ii) its temporal evolution follows physically consistent dynamics. To achieve this without requiring ground-truth physical annotations, PIWM employs weak distribution-based supervision that captures state uncertainty naturally arising from real-world sensing pipelines. The architecture integrates a VQ-based visual encoder, a transformer-based physical encoder, and a learnable dynamics model grounded in known physical equations. Across three case studies (Cart Pole, Lunar Lander, and Donkey Car), PIWM achieves accurate long-horizon prediction, recovers true system parameters, and significantly improves physical grounding over purely data-driven models. These results demonstrate the feasibility and advantages of learning physically interpretable world models directly from images under weak supervision.
\end{abstract}






\maketitle
\section{Introduction}

Accurate and robust trajectory prediction from high-dimensional sensor data, such as camera images, is a fundamental challenge for the safe operation of cyber-physical systems (CPS). A dominant paradigm for this task is to learn compact latent representations of the environment and evolve them over time. This idea underlies modern \textit{world models}~\cite{ha2018world}, which typically combine deep visual encoders with predictive temporal models such as RNNs or transformers~\cite{micheli2022transformers, hafner2023mastering, seo2023masked}. While these models achieve strong long-horizon prediction performance, their latent representations usually function as a ``black box,'' lacking a clear connection to the underlying physical state of the system.

Such a gap limits trustworthiness, controllability, and safety in CPS. In settings like autonomous driving or household robotics, physically meaningful latent states enable causal explanations (e.g., slowing down due to occlusions) and support high-assurance methodologies such as formal verification~\cite{hasan2015formal, katz2022verification} and run-time shielding~\cite{alshiekh2018safe, waga2022dynamic}. If predicted states can be mapped to interpretable quantities such as position or velocity, a safety monitor can detect unsafe future configurations (e.g., entering an occupied lane) and intervene accordingly. Furthermore, embedding physical structure into the latent space improves generalization by constraining predictions to physically plausible trajectories.

\begin{figure}[t] \centering \includegraphics[width=0.9\columnwidth]{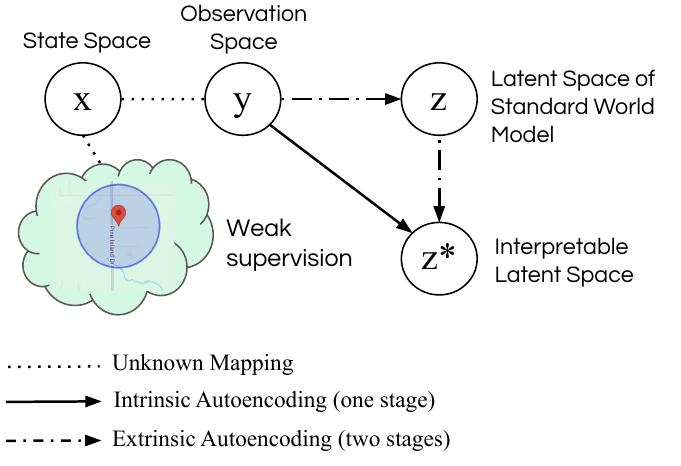} \caption{
Conceptual illustration of learning physically meaningful latent spaces from images.
High-dimensional observations $y$ may be encoded into a standard latent space $z$ or an interpretable latent space $z^*$.
Weak supervision vaguely relates observations to the underlying physical state $x$.
Existing approaches follow two paradigms: \emph{intrinsic} autoencoding (one-stage) and \emph{extrinsic} autoencoding (two-stage).
} \label{fig:1} \end{figure}

\looseness=-1
Existing attempts to learn physically interpretable latents from images generally fall into two paradigms, illustrated in Figure~\ref{fig:1}. \emph{Extrinsic} approaches first learn abstract visual latents and then map them to physical quantities via an additional model. \emph{Intrinsic} approaches instead attempt to encode physical structure directly within the image encoder, enforcing that the latent representation itself corresponds to meaningful physical variables. While effective in controlled settings~\cite{le2025pixie}, both paradigms typically \textit{require exact physical labels} during training or rely on object-centric decompositions~\cite{mosbach2025soldslotobjectcentriclatent}, which are difficult to reliably obtain from real-world CPS.

\looseness=-1
We aim to overcome this limitation by learning physical representations \emph{without exact physical supervision}. In many CPS, sensors such as GPS and radar naturally produce uncertain estimates in the form of distributions or confidence intervals rather than precise measurements. These distributional signals provide coarse yet informative constraints. For example, approximate position ranges or speed bounds that reflect the true physical state while tolerating noise and ambiguity. We leverage such distribution-based weak supervision to guide the encoding of high-dimensional images toward physically meaningful latent states. Combined with partially known system dynamics (which is also typical for CPS), this enables physically consistent multi-step prediction without requiring access to ground-truth states.

\looseness=-1
To this end, we propose the \textit{Physically Interpretable World Model (PIWM)}, a framework that learns latent states aligned with real physical quantities and constrains their temporal evolution using a structured, partially known dynamics model. In this work, \emph{physical interpretability} encompasses two complementary properties: (i) some latent dimensions correspond to meaningful physical variables, and (ii) latent evolution follows physically consistent dynamics. PIWM provides a general architectural template that combines image-based representation learning, physically interpretable latent spaces, and physics-informed dynamics.
Rather than prescribing a single architecture, PIWM defines a design space that can be instantiated with intrinsic or extrinsic encoders, and with continuous or discrete latent parameterizations.
Within this space, physical interpretability arises from aligning latent variables with weak supervision and constraining their temporal evolution through structured dynamics. Simultaneously, weak distributional supervision enforces alignment between latent states and physical quantities, eliminating reliance on the exact ground truth for training.

We conduct an extensive experimental study to answer a central question: \emph{How can weak supervision and latent quantization make world models physically interpretable?} We systematically examine continuous vs. discrete latent spaces as well as intrinsic vs. extrinsic designs. Across three case studies --- CartPole, Lunar Lander, and the DonkeyCar autonomous racing platform --- PIWM achieves superior physical grounding and long-horizon prediction accuracy compared to purely data-driven models and baselines. Notably, extrinsic architectures with quantized latent spaces consistently yield the most physically plausible and robust representations, highlighting the importance of decoupling visual encoding from physical interpretation. 

Thus, this paper makes three contributions:
\begin{enumerate}
    \item \textbf{A unified definition of physical interpretability.}
    We formalize physical interpretability for generative world models as two complementary properties:
    (i) latent representations correspond to physical quantities, and
    (ii) their temporal evolution obeys physically valid dynamics.

    \item \textbf{A novel weakly supervised framework for learning physically grounded latent states.}
    We introduce an architecture and training pipeline that aligns image-based latent states with physical variables using \emph{distribution-based weak supervision} instead of exact physical annotations. It also makes use of additional physical priors in the form of structural dynamics and latent quantization. 


    \item \textbf{A systematic empirical study of latent design choices.}
    Through extensive experiments on CartPole, Lunar Lander, and DonkeyCar, we analyze intrinsic vs.\ extrinsic architectures and continuous vs.\ discrete latents. The results reveal that \emph{extrinsic architectures with quantized latents} consistently achieve the most robust and interpretable representations, outperforming data-driven baselines in long-horizon prediction and physical parameter recovery.
\end{enumerate}

The remainder of this paper is organized as follows.
Section~\ref{sec:preliminaries} introduces the necessary background on autonomous systems, world models, and learned representations.
Section~\ref{sec:architecture} presents the proposed framework, including the representation learning module, the structured dynamics model, and the unified training procedure.
Section~\ref{sec:experiments} reports an extensive empirical study across three case studies, evaluating predictive performance, physical interpretability, and the effects of latent design choices. Section~\ref{sec:dis} discusses our findings, whereas Section~\ref{related} reviews the existing work.
Finally, Section~\ref{sec:conclusion} concludes the paper.

\section{Preliminaries}\label{sec:preliminaries}

\subsection{Autonomous system and world models}
\begin{definition}[Autonomous CPS]
An \emph{autonomous CPS} $s = (X, I, Y, A, \phi_{\theta}, g, h)$ models the evolution of an perception-based control system, where the state set $X$ defines the finite-dimensional space of physical states, the initial set $I \subset X$ specifies possible starting states, observation set $Y$ contains all possible observations, and the action set $A$ contains available control actions. The system dynamics $\phi_{\theta}: X \times A \times \Theta \rightarrow X$ governs state transitions under physical parameters $\theta \in \Theta$, the observation function $g: X \rightarrow Y$ maps states to observations, and the fixed controller $h: Y \rightarrow A$ selects actions based on observations.
\end{definition}

We consider an autonomous system with observation-based control, the states of which evolve as $x_{t+1} = \phi(x_t, a_t, \theta^*),$ where $\theta^*$ are the true (unknown) physical parameters. The known controller $h$ relies on observations $y$ from sensors (e.g., cameras). Such controllers can be trained by imitation learning~\cite[]{hussein2017imitation} or reinforcement learning~\cite[]{kaelbling1996reinforcement,lillicrap2015continuous}, and we assume that it has been done for our purposes.
Given an initial state \( x_0 \in I \), a \textit{state trajectory} is defined by iteratively applying the dynamics function \( \phi \), yielding a sequence \( (x_0, x_1, \ldots, x_{i}) \), where each state is \( x_{t+1} = \phi(x_t, a_t, \theta^*) \) and the action is generated by the controller based on the observation:  \( a_{t}  = h(y_t) = h(g(x_t)) \). Correspondingly, an \textit{observation trajectory} \( (y_0, y_1, \ldots, y_{t}) \) is generated by applying the observation function \( g \) to each state: \( y_t = g(x_t) \).

We consider the setting where the state is not directly observable, and to anticipate hazards and adapt, the system needs to forecast observations. This is done by combining a world model conditioned on past observations with an observation-based controller $h$, essentially forming an efficient online simulator of future observations. 

\begin{definition}[World Model]
A \emph{world model} $\mathcal{W} = (\mathcal{E}, f, \mathcal{D})$ predicts the future evolution of observations in an autonomous CPS $s$, where the encoder $\mathcal{E}: Y \rightarrow Z$ compresses high-dimensional observations into latent representations, the predictor $f: Z \times A \rightarrow Z$ forecasts future latents conditioned on actions, and the decoder $\mathcal{D}: Z \rightarrow Y$ reconstructs predicted observations. The actions $a_t$ are generated by a controller $h: Y \rightarrow A$ based on the current observation $y_t$, and together with the encoded latent $z_t = \mathcal{E}(y_t)$, are used to obtain the future latent $z_{t+1} = f(z_t, a_t)$. The predicted observation $\hat{y}_{t+1}$ is then obtained by decoding $z_{t+1}$ with $\mathcal{D}: \hat{y}_{t+1} = \mathcal{D}(z_{t+1}) $.
\end{definition}

While world models enable diverse tasks (e.g., controller training, monitoring, and planning), 
they are usually evaluated with \textit{predictive accuracy} --- how closely the predicted observations $\hat{y}_{t+1:t+n}$ match the true observations $y_{t+1:t+n}$. The typical metrics include mean squared error (MSE) and structural similarity index (SSIM). 
Unfortunately, these metrics are inherently limited: similar pixel-wise observations may correspond to very different underlying states. This means that the similarity (low MSE and high SSIM) between $\hat{y}_{t+1:t+n}$ and  $y_{t+1:t+n}$ does not guarantee similarity between the underlying physical states corresponding to the predicted observations and the true physical ones. Thus, the error in predicted observations is \textit{not physically interpretable}. This severely limits the predictions' utility for downstream CPS tasks like safety monitoring and planning. Instead, we turn our attention to the latent states. 

\subsection{Interpretable Latent Representations}


 \begin{figure*}[t]
	\centering     
\includegraphics[width=1.7\columnwidth]{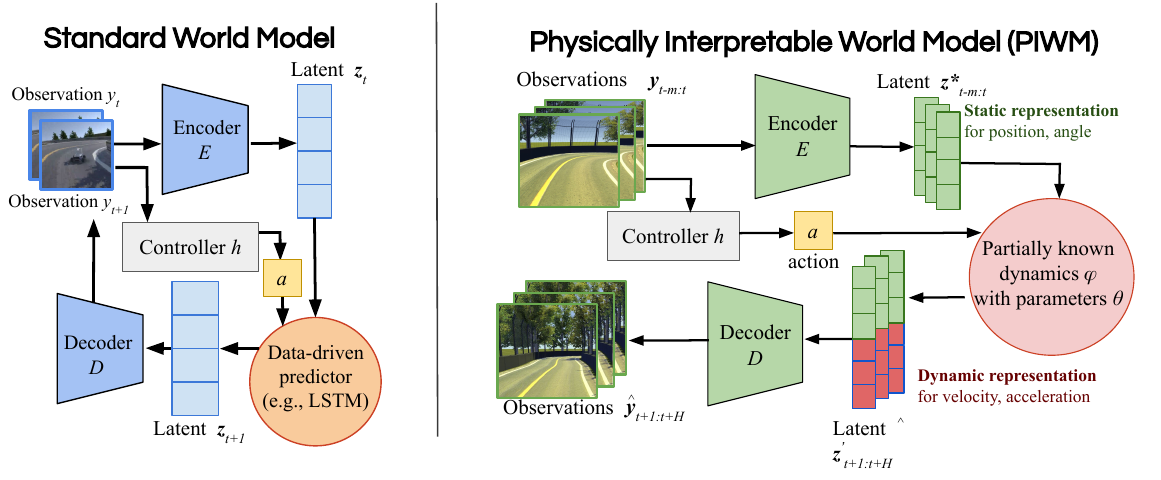}
\caption{Overview of world model architectures. {(Left)}  Standard world model uses an encoder-decoder structure with a data-driven predictor (e.g., LSTM) to model latent dynamics, where the latent representations lack physical meaning. {(Right)} The PIWM architecture learns a structured latent representation $z^*$ from images, uses a learnable dynamics model $\phi$ with physical priors to predict future latent states, and decodes them into future images.}	\label{fig:15}
 \end{figure*}

The autoencoder components $\mathcal{E}$ and $\mathcal{D}$ in world models are primarily based on Variational Autoencoders (VAEs), Vector-Quantized Variational Autoencoder (VQ-VAEs), or their variants. VAEs~\cite[]{kingma2013auto} encode high-dimensional inputs \( y \) into continuous latent vectors \( z = \mathcal{E}(y) \), which are then decoded back into reconstructed inputs \( \hat{y} = \mathcal{D}(z) \). A standard VAE assumes a simple prior distribution over the latent space, typically an isotropic Gaussian, which facilitates tractable inference but imposes a strong inductive bias.  A VQ-VAE~\cite[]{van2017neural} discretizes the latent space by mapping a high-dimensional observation \( y_t \) to a continuous latent vector \( z_t = \mathcal{E}(y_t) \), then quantizing it via a \textit{codebook} \( \{\mathbf{e}_k\}_{k=1}^K \) to obtain a discrete latent \textit{index} \( z_t^* = \arg\min_k \|z_t - \mathbf{e}_k\|_2^2 \) closest to the continuous latent $z_t$. The decoder then reconstructs the observation via \( \hat{y}_t = \mathcal{D}(\mathbf{e}_{z_t^*}) \). While discretization introduces structure and regularization, the learned codebook vectors \( \mathbf{e}_k \) themselves are typically unstructured and lack interpretation.  Therefore, merely discretizing the space is insufficient: an explicit mechanism is needed to align these latent representations with physical semantics.

\looseness=-1
Our setting assumes that general knowledge about the dynamics function \( \phi \), namely the structure of its equations. However, the dynamics parameters \( \theta \), such as the mass or friction coefficient, are unknown, making the system $s$ only partially specified. In practice, it is often hard to find the true parameters $\theta^*$ because true physical states $x$ cannot be measured precisely. However, it is typical to compute a \textit{distribution} $p(x)$ from high-dimensional observations. For instance, the robot's pose may be estimated from images as a range/distribution --- and serve as a weak supervisory signal.

Thus, to train a world model, we are given a dataset \( \mathcal{S} \) of trajectories, consisting of \( N \) sequences of length \( M \), where each sequence contains tuples of images, actions, and state distribution labels:
\begin{equation}
 \mathcal{S} = \Big\{ \big(y_{1:M}, a_{1:M}, p_{1:M}(x) \big) \Big\}_{1:N}
\end{equation}

The weak supervision is provided by state distributions \( p_t(x) \), for which we assume no analytical form is known. Instead, we can only draw a finite set of samples from them. This represents a challenging yet practical scenario, as it is a weaker assumption than knowing the distribution's type (e.g., Gaussian), its parameters (e.g., mean and variance), or a definitive interval. 

\looseness=-1
Our ultimate task is to learn a representation \( z^* \) that accurately approximates the system’s true, unobserved physical state \( x \) for prediction. We formalize this as two distinct but related problems:

\begin{definition}[Interpretable Representation Learning Problem]
Given a dataset \( \mathcal{S} \) and a controller \( h \), the objective is to \emph{learn a state representation} \( z^* = \mathcal{E}(y) \) that minimizes the mean squared error to the true physical state \( x \):
\begin{equation}
\min_{\mathcal{E}} \ \mathbb{E} \big[ \| x - \mathcal{E}(y) \|_2^2 \big]
\end{equation}
\label{def:repprob}
\end{definition}

Next, the prediction problem is to forecast the evolution of these interpretable representations over time:

\begin{definition}[Prediction Problem for Interpretable Representations]
\looseness=-1
Given a dataset \( \mathcal{S} \), a dynamics function \( \phi \) with unknown parameters, and a controller \( h \), the objective is to \emph{train a predictor} that maps a history of representations to a future representation, \( \hat{z}^*_{t+H} = \mathcal{P}(z^*_{t-m:t}) \), by minimizing the future state prediction error:
\begin{equation}
\min_{\mathcal{P}} \ \mathbb{E} \big[ \| x_{t+H} - \hat{z}^*_{t+H} \|_2^2 \big]
\end{equation}
\label{def:predprob}
\end{definition}

The objectives in Definitions \ref{def:repprob} and \ref{def:predprob} are formulated with respect to the true physical state \( x \), which serves as the ultimate ground truth for evaluating physical interpretability. However, since \( x \) is not accessible during training, these objectives cannot be optimized directly. Our proposed method, detailed in Section~\ref{sec:architecture}, addresses this challenge by constructing a tractable surrogate objective that leverages the weak supervision by sampling state distributions \( p(x) \). Throughout this paper, we use calligraphic letters (e.g., $\mathcal{E}$, $\mathcal{D}$, $\mathcal{S}$) to denote functions and datasets for clarity. During training, we optimize model parameters $w$ using mini-batch gradient descent with batch size $B$, learning rate $\eta$, and train for $n_{\text{epochs}}$ epochs.



\paragraph{Noise Model.}
Since the true physical state $x$ is not directly observable, the weak supervision available during training consists of samples drawn from an unknown distribution $p(x)$ obtained from noisy sensor data.
We assume only that $p(x)$ provides a \emph{mean-proximal} estimate of each state dimension. 
For the $i$-th state component, let $\mathcal{X}_i$ denote the valid range size of that dimension. Then the weak supervision is such that the true value $x_i$ lies within an  interval of relative width~$\delta$ centered at the distributional mean:
\[
x_i \in 
\Big[
\mathbb{E}[p(x)] - \tfrac{1}{2}\delta\,|\mathcal{X}_i|,\;
\mathbb{E}[p(x)] + \tfrac{1}{2}\delta\,|\mathcal{X}_i|
\Big]
\]
Here, hyperparameter $\delta\in(0,1)$ quantifies the weakness of the supervision.  
This interval characterizes the feasible set from which supervision samples are drawn. In practice, we synthesize biased supervision noise to satisfy this assumption in Section~\ref{sec:setup}. It is important to distinguish our weak supervision from standard noisy-label supervised learning. First, PIWM never accesses the ground-truth state $x$ directly; it operates only on samples drawn from an unknown distribution $p(x)$. Second, no assumption is made about the analytical form or parameters of this distribution (e.g., we do not assume a Gaussian with known variance). Third, the interpretability objective aligns the latent state with the empirical mean of the sample set, not with a single corrupted regression target. This formulation reflects a more challenging and realistic scenario, as the model must infer the underlying state without any parametric description of the sensing noise.

We emphasize that our supervision is strictly weaker than noisy-label regression: the model never observes even a corrupted version of the true state, but only a finite sample set from an unknown distribution whose form and parameters are inaccessible.

\paragraph{Other assumptions.} 
\looseness=-1
Our method is designed under several standard and widely adopted assumptions for image-based CPS. First, we assume partial observability from images: the subset of physical variables we aim to recover is identifiable from a sequence of visual observations. Second, we adopt the Markovian system assumption, consistent with most world-model and control formulations: the next state depends only on the current state and action. Third, we assume that a partially known dynamics model provides a reasonable approximation of reality, but it does not need to be exact.


\section{PIWM Architecture}\label{sec:architecture}

We introduce the \textit{Physically Interpretable World Model}, a flexible prediction architecture with physically-grounded representations of high-dimensional observations. It consists of two core components: (1) a {physically interpretable autoencoder} responsible for learning the state representation, and (2) a {learnable dynamics model} that predicts the evolution of this representation. A high-level overview of the proposed PIWM architecture is shown in Fig.~\ref{fig:15}, which contrasts it with a standard world model and highlights the structured, interpretable latent state. 

In our formulation, physical interpretability has two components. 
First, the learned physical latent variables $z^*$ correspond to meaningful physical quantities such as position, velocity, or other environment-specific state variables $x$. 
Second, the dynamics model $\phi$ learns to follow physically grounded evolution, ensuring that the temporal behavior of $z^*$ aligns with the underlying laws of motion. 
Our goal is to achieve both forms of interpretability directly from high-dimensional image observations under weak supervision. In our formulation, the physical state $x$ refers to a task-relevant, modeled subset of the full world state, namely the variables captured by the known dynamics equations $\phi$, such as position, velocity, and angle. Factors not captured by this subset (e.g., background motion, unmodeled friction, or moving objects) are not explicitly constrained by the symbolic dynamics and are instead implicitly absorbed into the visual components of the latent representation in both intrinsic and extrinsic variants. The decoder must reconstruct the full observation from the combined physical and visual latent components, which incentivizes the visual component to capture these residual factors.

\subsection{Learning Interpretable Representations}

The primary goal of the autoencoder is to map a high-dimensional observation \(y\) to a low-dimensional interpretable latent state \(z^*\). This representation must be explicitly aligned with the true physical state \(x\). Per Section~\ref{sec:preliminaries}, we cannot access \(x\) or the analytical form of its supervisory distribution \(p(x)\). Instead, we are given access to a set of \(L\) state proxy samples, \(\Xi = \{\xi^{(l)}\}_{l=1}^L\), drawn from \(p(x)\). To leverage these proxy samples \(\Xi\) for training, we formulate a general {interpretability loss}, \(\mathcal{L}_{\text{interp}}\), which measures the discrepancy between a predicted physically interpretable state \(z_p^*\) and the sample set \(\Xi\). In our experiments, we primarily use a Mean Squared Error (MSE) formulation that penalizes the distance to the empirical mean of the samples:
\begin{equation}
\mathcal{L}_{\text{interp}}(z_p^*, \Xi) = \| z_p^* - \hat{\mu}_{\xi} \|_2^2, \quad \text{where} \quad \hat{\mu}_{\xi} = \frac{1}{L}\sum_{l=1}^L \xi^{(l)}
\label{eq:interp_loss}
\end{equation}
An alternative could be to use the Kullback-Leibler (KL) Divergence against a Gaussian fitted to the samples. This serves as the mechanism for enforcing physical grounding throughout our models.

\looseness=-1
A central challenge in learning representation \(z^*\) is managing two competing objectives: reconstructing high-dimensional observations versus aligning the latent space with low-dimensional physical states. We consider two approaches, previously highlighted in Fig.~\ref{fig:1}. The \emph{Intrinsic} approach attempts to achieve both objectives simultaneously with a single, end-to-end encoder. While potentially more efficient, this forces one network to both capture fine-grained visual details (for reconstruction) and ignore those same details to extract the underlying physical state --- a difficult learning task. In contrast, the \emph{Extrinsic} approach follows a two-stage process: a vision autoencoder first learns an intermediate representation focused on reconstruction, and then a second, physical encoder extracts the interpretable state from it. This modularity may stabilize training but risks information loss in the intermediate step. Given the fundamental trade-offs, we will systematically investigate both approaches.

\looseness=-1
Orthogonal to this architectural choice, a second key design decision is the nature of the latent space \(Z^*\). \emph{Continuous spaces} can represent high-fidelity physical quantities but lack a built-in organizational prior, requiring explicit regularization for disentanglement. Conversely, \emph{discrete spaces} enforce regularity by design through their finite codebook but lose precision due to quantization error.
Below, we detail the combinations of intrinsic/extrinsic and continuous/discrete approaches. Note that the individual components, namely VAE/VQ-VAE encoders, $\beta$-VAE regularization, and second-order dynamics, are standard building blocks; the novelty of PIWM lies in their systematic integration with distribution-based weak supervision and the resulting design-space analysis.

\subsubsection{Intrinsic Autoencoding}

\looseness=-1
This approach employs a single, end-to-end encoder \(\mathcal{E}: Y \to Z^*\) that directly maps an observation \(y\) to the final interpretable latent state \(z^*\). This unified representation must disentangle visual features from physical semantics, a known challenge where information may leak between different physical attributes and hinder the desired learning \cite[]{peper2025four}.
For the case of \textit{continuous} latent space, the encoder outputs the parameters for a posterior distribution \(q(z^*\mid y)\) over the latent space \(Z^*\). Sampled from this distribution, a latent vector is then partitioned in a fixed manner:  \(z^* = [z_p^*, z_v^*]\), where \(z_p^*\) is the physically interpretable part and \(z_v^*\) captures the remaining visual information necessary for reconstruction. The full objective follows the \(\beta\)-VAE formulation, with the interpretability loss applied only to \(z_p^*\) and KL loss applied to $z_v^*$:
\begin{equation}
\begin{split}
    \mathcal{L}_{\text{intrinsic-cont}} = ~&\mathcal{L}_{\text{recon}}(y, \hat{y}) + \lambda_{\text{interp}}\mathcal{L}_{\text{interp}}(z_p^*, \Xi) \\
    &+ \beta D_{\text{KL}}\big(q(z_v^*|y) \,\|\, \mathcal{N}(0, I)\big)
\end{split}
\end{equation}

\looseness=-1
For the case of \textit{discrete} latent space, we structure the codebook to be interpretable. Each vector \(\mathbf{e}_k\) is partitioned as \(\mathbf{e}_k = [\mathbf{e}_k^p, \mathbf{e}_k^v]\).  Only the visual part \(\mathbf{e}_k^v\) is a typical learnable VQ-VAE codebook vector.  The physical part \(\mathbf{e}_k^p\) is a constant vector representing a specific point in a discretized grid of physical values (e.g., specific positions). Then, the interpretable state is computed as the average of the physical portions of the codebook vectors, \( z_p^* = \frac{1}{|I|} \sum_{i \in I} \mathbf{e}_i^p \). The full objective combines the standard VQ loss with our interpretability loss:
\begin{equation}
    \mathcal{L}_{\text{intrinsic-disc}} = \mathcal{L}_{\text{VQ}}(y, \hat{y}) + \lambda_{\text{interp}}\mathcal{L}_{\text{interp}}(z_p^*, \Xi),
\end{equation}
where \(\mathcal{L}_{\text{VQ}}\) includes reconstruction, codebook, and commitment losses:
\begin{equation}
    \mathcal{L}_{\text{VQ}} = \big\| y - \hat{y} \big\|_2^2 + \big\| \text{sg}[z_{\text{cont}}] - z_q \big\|_2^2 + \beta \big\| z_{\text{cont}} - \text{sg}[z_q] \big\|_2^2
\end{equation}

Here, \(z_{\text{cont}}\) is the continuous output of the encoder \(\mathcal{E}\), and \(z_q\) is its nearest vector from the codebook. The \(\text{sg}[\cdot]\) is the stop-gradient operator, which ensures that gradients are routed correctly for the codebook loss (updating \(z_q\)) and the commitment loss (updating \(z_{\text{cont}}\)). The hyperparameter \(\beta\) weights this commitment loss, controlling how strongly the encoder's output is encouraged to match the chosen codebook vector.

\subsubsection{Extrinsic Autoencoding}

This approach utilizes a two-stage training process to decouple perception from interpretation.
First, a general-purpose vision autoencoder (\(\mathcal{E}_v, \mathcal{D}_v\)) is trained to map an observation \(y\) to an intermediate latent vector \(z = \mathcal{E}_v(y)\). This stage is trained with a standard objective, independent of physical supervision.
For the \textit{continuous} case, this is a \(\beta\)-VAE trained to minimize:
\begin{equation}
    \mathcal{L}_{\text{vision-cont}} = \mathcal{L}_{\text{recon}}(y, \hat{y}) + \beta D_{\text{KL}}\big(q(z \mid y) \,\|\, \mathcal{N}(0, I)\big)
\end{equation}
For the \textit{discrete} case, a VQ-VAE is trained to minimize the standard VQ loss \(\mathcal{L}_{\text{vision-disc}} = \mathcal{L}_{\text{VQ}}(y, \hat{y})\).

After the first stage, the vision encoder \(\mathcal{E}_v\) is frozen. The second stage trains a separate, auxiliary {physical autoencoder} (\(\mathcal{E}_p, \mathcal{D}_p\)) to map the intermediate representation \(z = \mathcal{E}_v(y)\) to a final, purely physical representation \(z^* = \mathcal{E}_p(z)\). The training objective for this stage is:
\begin{equation}
    \mathcal{L}_{\text{physical}} = \lambda_{\text{interp}}\mathcal{L}_{\text{interp}}(z^*, \Xi) + \lambda_{\text{latent}}\mathcal{L}_{\text{recon}}\big(z,  \mathcal{D}_p(z^*))
\end{equation}

This same loss \(\mathcal{L}_{\text{physical}}\) is used for both continuous and discrete intermediate representation \(z\). Architectural diagrams of the extrinsic-discrete pipeline are provided in Figure~\ref{fig:extrinsic-vq-appendix} in the Appendix.

\begin{algorithm}[t]
\caption{
Training pipeline for PIWM. Models are trained in separate stages, freezing the
parameters of earlier stages before optimizing the next one.
}
\label{alg:piwm}
\begin{algorithmic}[1]

\Require Training set $\mathcal{S}$, batch size $B$, weights $\{\lambda_i\}$

\State \textbf{// Stage 1: Train the vision autoencoder (extrinsic only)}
\If{architecture = extrinsic}
    \For{epoch $= 1$ to $n_{\text{vision}}$}
      \For{each batch of $(y_t)$}
        \State $z_t \gets \mathcal{E}_v(y_t)$
        \If{latent = discrete}
            \State $z_t \gets \text{Quantize}(z_t)$
        \EndIf
        \State $\hat{y}_t \gets \mathcal{D}_v(z_t)$
        \State $\mathcal{L}_{\text{vision}} \gets 
               \mathcal{L}_{\text{rec}}(y_t, \hat{y}_t) 
               + \lambda_3 \mathcal{L}_{\text{reg}}$
        \State Update $(\mathcal{E}_v, \mathcal{D}_v)$ using 
               $\nabla \mathcal{L}_{\text{vision}}$
      \EndFor
    \EndFor
    \State Freeze parameters of $\mathcal{E}_v$
\EndIf

\State \textbf{// Stage 2: Train the physical encoder-decoder}
\For{epoch $= 1$ to $n_{\text{physical}}$}
  \For{each batch of $(y_t, \Xi_t)$}
    \If{architecture = intrinsic}
        \State $z^*_t \gets \mathcal{E}(y_t)$
    \Else
        \State $z_t \gets \mathcal{E}_v(y_t)$
        \If{latent = discrete}
            \State $z_t \gets \text{Quantize}(z_t)$
        \EndIf
        \State $z^*_t \gets \mathcal{E}_p(z_t)$
    \EndIf
    \State Compute interpretability loss 
           $\mathcal{L}_{\text{interp}}(z^*_t, \Xi_t)$
    \If{architecture = extrinsic}
        \State $\hat{z}_t \gets \mathcal{D}_p(z^*_t)$
        \State $\mathcal{L}_{\text{latent-rec}} \gets 
               \|z_t - \hat{z}_t\|^2$
    \Else
        \State $\mathcal{L}_{\text{latent-rec}} \gets 0$
    \EndIf
    \State $\mathcal{L}_{\text{phys}} \gets 
           \lambda_1 \mathcal{L}_{\text{interp}}
           + \lambda_2 \mathcal{L}_{\text{latent-rec}}$
\State Update physical encoder-decoder using $\nabla \mathcal{L}_{\text{phys}}$
\EndFor
\EndFor
\State Freeze $\mathcal{E}$ (intrinsic) or $(\mathcal{E}_p, \mathcal{D}_p)$ (extrinsic)

\State \textbf{// Stage 3: Train the dynamics model}
\For{epoch $= 1$ to $n_{\text{dyn}}$}
  \For{each batch of $(y_{t:t+2}, a_t, \Xi_{t:t+2})$}
    \State Encode states to obtain $z^*_t$, $z^*_{t+1}$
    \State Predict next physical state 
           $\hat{z}^*_{t+2} \gets \phi_\theta(z^*_t, z^*_{t+1}, a_t)$
    \State Compute dynamics loss 
           $\mathcal{L}_{\text{dyn}}(\hat{z}^*_{t+2}, \Xi_{t+2})$
    \State Update $\theta$ using $\nabla \mathcal{L}_{\text{dyn}}$
  \EndFor
\EndFor

\State \Return trained model parameters
\end{algorithmic}
\end{algorithm}

\subsection{Learnable Dynamics Model}


The second core component of our PIWM is a latent dynamics model \(\phi\) that predicts the temporal evolution of the physically interpretable state \(z^*\). Rather than using black-box sequence models, our prediction is based on known dynamics equations, \(\phi(z^*_t, a_t, \theta)\), where the form of \(\phi\) (e.g., kinematics) is fixed and only its parameters \(\theta\) are learnable. This allows our model to reflect the underlying physical laws while adapting to unknown system parameters $\theta$ such as friction and mass.

The dynamics model is responsible for estimating physical quantities that depend on a history of states, such as velocity, in order to predict the system's evolution. 
To do this, the model is initialized with a short window of consecutive representations, \( (z^*_t, z^*_{t+1}) \), produced by the encoder from observations. 
These two states, along with the control action \(a_{t+1}\) that causes the transition from state \(t+1\) to \(t+2\), are used by the learnable dynamics model \(\phi\) to predict the next state: \( \hat{z}^*_{t+2} = \phi(z^*_t, z^*_{t+1}, a_{t+1}, \theta) \). 
By taking two consecutive states as input, the model can internally compute velocity and other time-derivative quantities necessary for an accurate physical prediction. This method can be extended to higher-order derivatives with more consecutive inputs.
After this initialization phase, the model can operate recursively for multi-step rollouts, taking its own prediction from the previous step as input to generate a future trajectory. 
This enables efficient, long-horizon forecasting without needing a sequence of observations at every step.

\looseness=-1
The parameters \(\theta\) of the dynamics model \(\phi\) are learned by minimizing a dynamics loss, \(\mathcal{L}_{\text{dyn}}\). This objective ensures that the predicted state \(\hat{z}^*_{t+H}\), generated by recursively applying the dynamics model \(\phi(\cdot, \cdot, \theta)\), aligns with the weak supervision available for that future time step. We use a Mean Squared Error (MSE) loss, which compares the prediction against the empirical mean of the proxy labels \(\Xi_{t+H}\). The loss is defined as a function of the parameters \(\theta\):
\begin{equation}
    \mathcal{L}_{\text{dyn}}(\theta) = \| \hat{z}^*_{t+H} -  \hat{\mu}_{\xi_{t+H}} \|_2^2,
\end{equation}
where $\hat{z}^*_{t+H}$ is the state predicted by the dynamics model parameterized by $\theta$, and $\hat{\mu}_{\xi_{t+H}} = \frac{1}{L}\sum_{l=1}^L \xi^{(l)}_{t+H}$ is the empirical mean of the $L$ proxy samples for the future state.

To optimize the learnable parameters $\theta$, we backpropagate 
through the structured \emph{second-order} dynamics model used in Algorithm~\ref{alg:piwm}.
Since each predicted state $\hat{z}^{*}_{t+H}$ is obtained by recursively applying
\begin{equation}
\hat{z}^{*}_{t+2} = \phi(z^{*}_{t},\, z^{*}_{t+1},\, a_{t+1};\, \theta),
\end{equation}
it depends on \emph{two} previous latent states, consistent with the model definition.
The gradient of the multi-step dynamics loss is:
\begin{equation}
\frac{\partial \mathcal{L}_{\mathrm{dyn}}}{\partial \theta}
=
\frac{\partial \mathcal{L}_{\mathrm{dyn}}}{\partial \hat{z}^{*}_{t+H}}
\cdot
\frac{\partial \hat{z}^{*}_{t+H}}{\partial \theta}
\end{equation}

The recursive derivative for $\hat{z}^{*}_{t+H}$ expands as:
\begin{equation}
\frac{\partial \hat{z}^{*}_{t+H}}{\partial \theta}
=
\frac{\partial \phi}{\partial \theta}
+
\frac{\partial \phi}{\partial z^{*}_{t+H-1}}
\cdot
\frac{\partial z^{*}_{t+H-1}}{\partial \theta}
+
\frac{\partial \phi}{\partial z^{*}_{t+H-2}}
\cdot
\frac{\partial z^{*}_{t+H-2}}{\partial \theta},
\end{equation}
reflecting the fact that the dynamics model $\phi$ takes these inputs:
$$(z^{*}_{t+H-2}, z^{*}_{t+H-1}, a_{t+H-1})$$

This recursion enables end-to-end learning of both the interpretable 
latents and the dynamics parameters, ensuring that the learned physical 
parameters remain consistent with observed trajectories. During training, the dynamics model is optimized after the representation modules have been trained and frozen.

\begin{figure*}[t]
	\centering         \includegraphics[width=1.87\columnwidth]{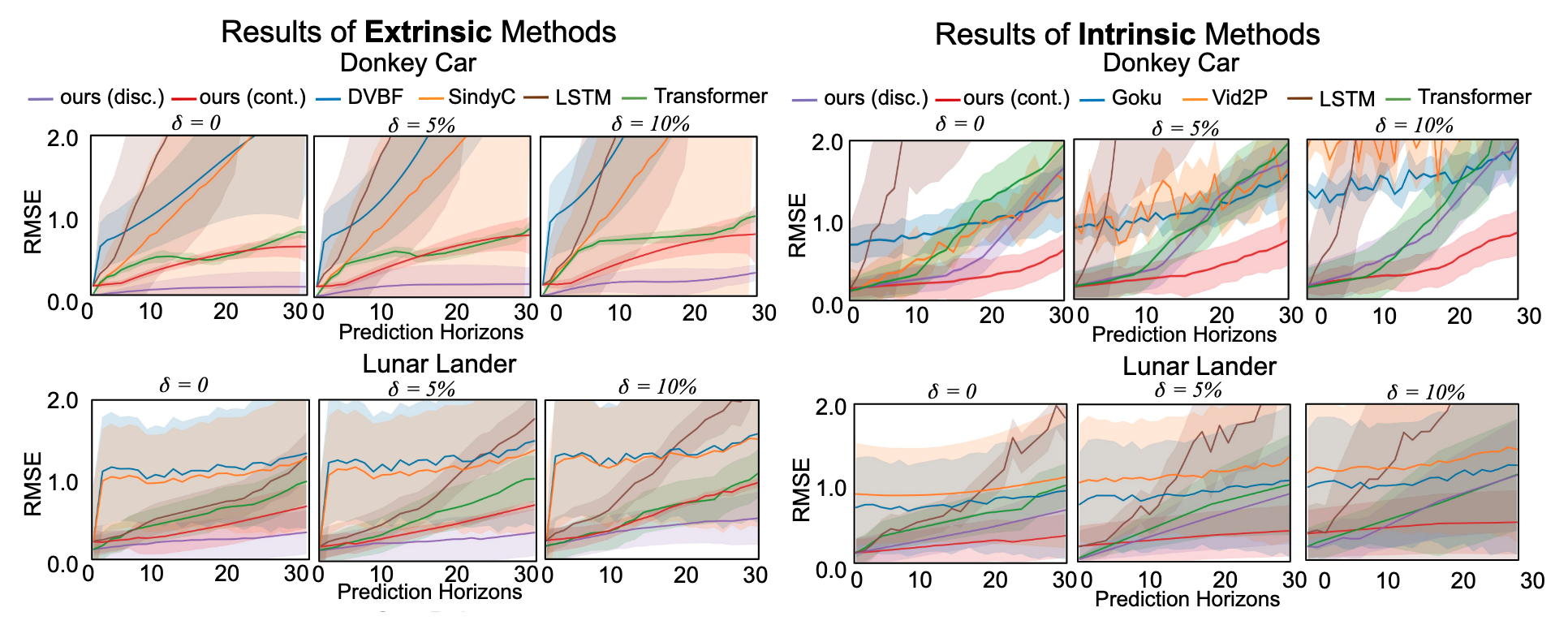}
\caption{Prediction performance. The Root Mean Square Error (RMSE) of our PIWM variants (extrinsic methods, left; intrinsic methods, right) is compared against baselines over a 30-step prediction horizon in the Donkey Car and Lunar Lander across varying levels of weak supervision ($\delta$).}	\label{fig:2}

 \end{figure*}



\subsection{Training Procedure}

\looseness=-1
PIWM is not trained end-to-end for all variants. 
Intrinsic and models are trained in two and three stages: the vision autoencoder is trained first and then frozen; the physical encoder-decoder is trained next and then frozen (only for the extrinsic one); finally, the dynamics model is trained with both encoders fixed. 
Algorithm~\ref{alg:piwm} summarizes the staged training procedure. The total loss combines $\mathcal{L}_{\text{rec}}$, $\mathcal{L}_{\text{interp}}$, $\mathcal{L}_{\text{dyn}}$, and $\mathcal{L}_{\text{reg}}$; only the relevant components are active in each stage.

\section{Experimental Evaluation}\label{sec:experiments}
To validate our PIWM approach, we experiment on three environments: CartPole, Lunar Lander, and the DonkeyCar autonomous racing platform~\cite[]{brockman2016openai,viitala2021learning}. These environments differ in the observation dimensionality, action space, and underlying dynamics.


\subsection{Experimental Setup}\label{sec:setup}

For each environment, we collect a dataset of 60,000 trajectories, each with at least 50 time steps. To ensure diverse state space coverage, trajectories are generated by executing both random actions and those generated by well-trained neural controllers. 

The weak supervision labels used in all experiments instantiate the 
uniform-distribution noise model described in Sec.~\ref{sec:preliminaries}.  
In practice, this means that the supervisory distribution $p(x)$ for
each state dimension is implemented as a biased uniform distribution over a 
bounded interval of relative width~$\delta$.  For each ground-truth 
state component $x_i$, let $\mathcal{X}_i$ denote the valid 
range size of that dimension.  We first generate a randomly shifted 
center:
\[
\tilde{\mu}_i 
= x_i + \Delta_i,
\qquad 
\Delta_i \sim \mathrm{Unif}\!\left[
    -\tfrac{1}{2}\delta \,|\mathcal{X}_i|,\;
    \tfrac{1}{2}\delta \,|\mathcal{X}_i|
\right].
\]
We then define the supervisory distribution itself as a uniform
distribution supported on an interval of the same width:
\[
p_i(x) = \mathrm{Unif}\!\left[
\tilde{\mu}_i - \tfrac{1}{2}\delta\,|\mathcal{X}_i|,\;
\tilde{\mu}_i + \tfrac{1}{2}\delta\,|\mathcal{X}_i|
\right].
\]
At each time step, we draw $50$ samples from this distribution to form 
the proxy supervision set $\Xi$ used in training.


Our architectures and training follow standard practices in world-model and VQ-VAE training. The vision encoder $\mathcal{E}_v$ is implemented with two convolutional layers with channel projection to a 64-dimensional latent space. Discrete variants employ a 512-entry codebook with commitment loss weight $\beta = 0.25$. The physical encoder $\mathcal{E}_p$ is a 2-layer Transformer (4 heads, 512 feedforward dimension) with mean pooling and linear projection to physical states. The decoder $\mathcal{D}_p$ mirrors this architecture. 
We apply early stopping based on validation loss to prevent overfitting and to ensure stable convergence across environments. 
In addition, we use a cosine learning-rate decay schedule to improve long-horizon rollout stability for the second-order latent dynamics model. We performed a small-scale sweep over candidate learning rates and regularization magnitudes and selected the configuration that yielded stable reconstruction loss, non-diverging rollouts, and consistent convergence across datasets.

\begin{figure}[t]
	\centering         \includegraphics[width=\columnwidth]{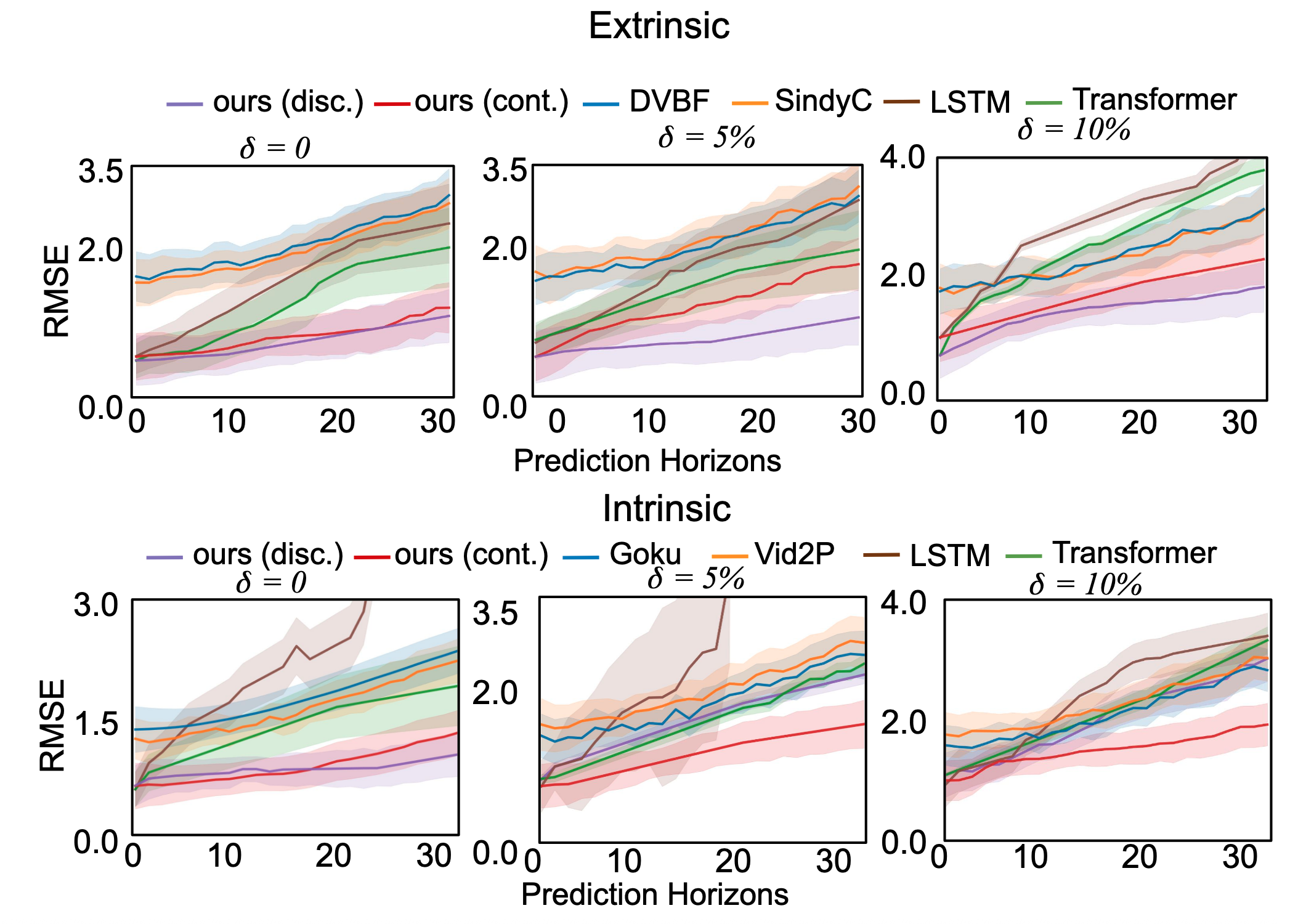}
\caption{Prediction performance of CartPole. The Root Mean Square Error (RMSE) of our PIWM variants (extrinsic methods, upper; intrinsic methods, lower) is compared against baselines over a 30-step prediction horizon in the CartPole across varying levels of weak supervision ($\delta$).}	\label{fig:para}
 \end{figure}


\looseness=-1
We evaluate our PIWM variants (Intrinsic/Extrinsic, Continuous/Discrete) against a suite of strong baselines under 5-fold cross-validation. We first include data-driven sequence models, an \textit{LSTM} and a \textit{Transformer}, which serve as non-physical benchmarks. Our primary comparisons are to state-of-the-art models in two categories. For the {intrinsic} approach, we compare against \textit{Vid2Para}~\cite[]{asenov2019vid2param} and \textit{GokuNet}~\cite[]{linial_generative_2021}. For the {extrinsic} approach, we evaluate against \textit{DVBF}~\cite[]{karl2016deep}. To specifically isolate and compare the performance of the latent dynamics predictors, we also include \textit{SindyC}~\cite[]{BRUNTON2016710} --- a classic state-based method for dynamics discovery. For a fair comparison, both the DVBF and SindyC dynamics models operate on the interpretable latent representations produced by our continuous autoencoder. All models are configured to have a comparable parameter count to compare architectural efficacy rather than model capacity. Furthermore, all baselines are implemented following their original publications or official codebases and trained under identical settings: the same dataset, rollout horizon, optimizer (Adam), learning rate schedule, and early stopping criterion based on validation loss. No additional hyperparameter tuning was performed exclusively for PIWM; all configuration choices were made on a held-out validation split shared across all models. More details can be found in the Appendix.

The controller used in all experiments is pretrained and kept fixed throughout the world model's training. Specifically, the controller is trained separately using ground-truth states under a standard supervised imitation-learning objective. The controller remains frozen and serves only as an action generator, without any joint optimization with the image encoder, the physical encoder, or the dynamics model. Thus, all models including baselines are trained using free-rollout prediction without teacher forcing. This setup reflects our target use case: free-running long-horizon prediction for safety monitoring and planning, where model outputs are recursively fed back as inputs at test time. In contrast, teacher forcing can mask compounding errors during training, leading to unstable rollouts under deployment conditions.

  \begin{figure*}[t]
  \centering
  \includegraphics[width=0.83\textwidth]{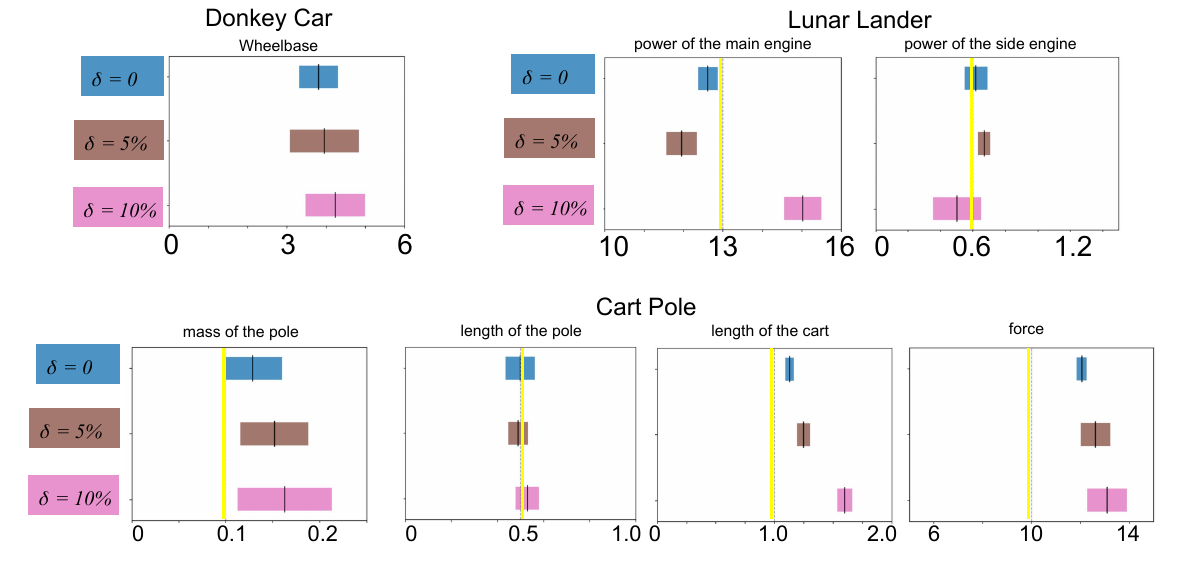} 
\caption{
Learned physical parameters vs.\ ground truth. 
For Cart Pole and Lunar Lander, the parameters learned by our model 
(colored bars) are compared against the known ground-truth values 
(yellow lines) under varying noise levels ($\delta$). 
For the Donkey~Car, we use only an approximate bicycle model 
and the true physical parameters are unknown; therefore, no ground-truth 
reference is shown.
}
\label{fig:3}
 \end{figure*}

\subsection{Predictive Performance}

\paragraph{Extrinsic vs.\ intrinsic performance.}
We first evaluate the primary task of long-horizon trajectory prediction. Figure~\ref{fig:2} shows the root mean square error (RMSE) for 30-step future state prediction in the challenging Donkey Car environment, comparing both extrinsic and intrinsic methods across different levels of supervision noise, \(\delta\).
The results for extrinsic methods (Fig.~\ref{fig:2}) show a clear performance hierarchy. Our PIWM variants consistently outperform all baselines. The {quantized extrinsic model (purple line)} achieves the lowest and most stable prediction error, maintaining accuracy even as the prediction horizon extends. Our continuous extrinsic model (red line) is the second-best performer. Both significantly surpass the extrinsic baselines (DVBF, SindyC) and the purely data-driven models (LSTM, Transformer), whose errors escalate rapidly.

A more nuanced picture emerges for the intrinsic methods (Fig.~\ref{fig:2}, right). Our PIWM models again demonstrate a clear advantage over the Vid2Para and GokuNet baselines. Strikingly, our continuous intrinsic model (red line) achieves a predictive accuracy that is highly competitive with, and in some cases even surpasses, our top-performing extrinsic models. This suggests that a well-regularized, end-to-end continuous architecture can be highly effective. In contrast, the quantized intrinsic model (purple line) exhibits less stability and higher error in this configuration, indicating that the optimization challenge of aligning a discrete codebook within a single, unified encoder is considerable. Nevertheless, both of our intrinsic variants outperform the baseline models, confirming the overall benefit of our training methodology.

\paragraph{Continuous vs.\ discrete latent spaces.}
A key insight from these results is that {decoupling visual perception from physical state inference (the extrinsic approach) is a critical design choice for achieving robust, long-term prediction.} An important observation from our results is the different strengths exhibited by continuous and discrete latent parameterizations across the intrinsic and extrinsic settings. In the intrinsic (one-stage) architecture, the encoder must jointly learn visual abstractions and physically structured representations. This joint optimization benefits from the flexibility and smooth gradients of a continuous latent space, which can more easily adapt to the coupled visual-physical objectives. In contrast, the extrinsic (two-stage) architecture decouples visual encoding from physical interpretation. In this setting, the discrete latent space acts as a strong regularizer: the quantization suppresses visual noise, stabilizes the mapping to physical states, and improves generalization of the dynamics model. Consequently, the continuous variant performs better under the intrinsic setup, while the discrete variant achieves superior performance under the extrinsic setup. 


\paragraph{Takeaway.} The best latent parameterization depends on whether physical structure is learned jointly with perception or inferred from a pre-learned visual representation.
Furthermore, across both architectures, the quantized (discrete) latent space provides a powerful regularization effect, leading to more stable predictions than the continuous alternative, especially under noisy supervision.

\subsection{Physicality of Learned Representations}

Strong predictive performance should stem from accurate underlying states and dynamics. We validated this observation by evaluating the static encoding quality and the model's ability to recover the true physical parameters of the simulation.

We also examined static encoding quality by measuring how accurately each model infers the physical state from a single observation. The results strongly correlate with the predictive findings: the quantized extrinsic PIWM achieves the highest encoding accuracy, confirming that its superior representation is the foundation of its predictive power. The intrinsic continuous models, while competitive, exhibit higher encoding error, consistent with their weaker predictive performance.

Finally, we assessed whether the dynamics model can learn the true physical parameters (e.g., car length, pole mass) within the learned latent space. As reported in Figure~\ref{fig:3}, PIWM successfully recovers the ground-truth parameters with low relative error across all environments. This provides direct evidence that our framework learns a genuinely interpretable representation that is not merely correlated with the physics --- but is structured in a physically grounded way. We emphasize that the DonkeyCar study is not designed to validate precise parameter recovery, since ground-truth physical parameters are unavailable, but rather to evaluate rollout stability and physical plausibility under approximate kinematic priors in a realistic robotic setting with visual sensing and control.

\section{Discussion and Future Work}\label{sec:dis}

\begin{figure*}[t]
	\centering         \includegraphics[width=1.95\columnwidth]{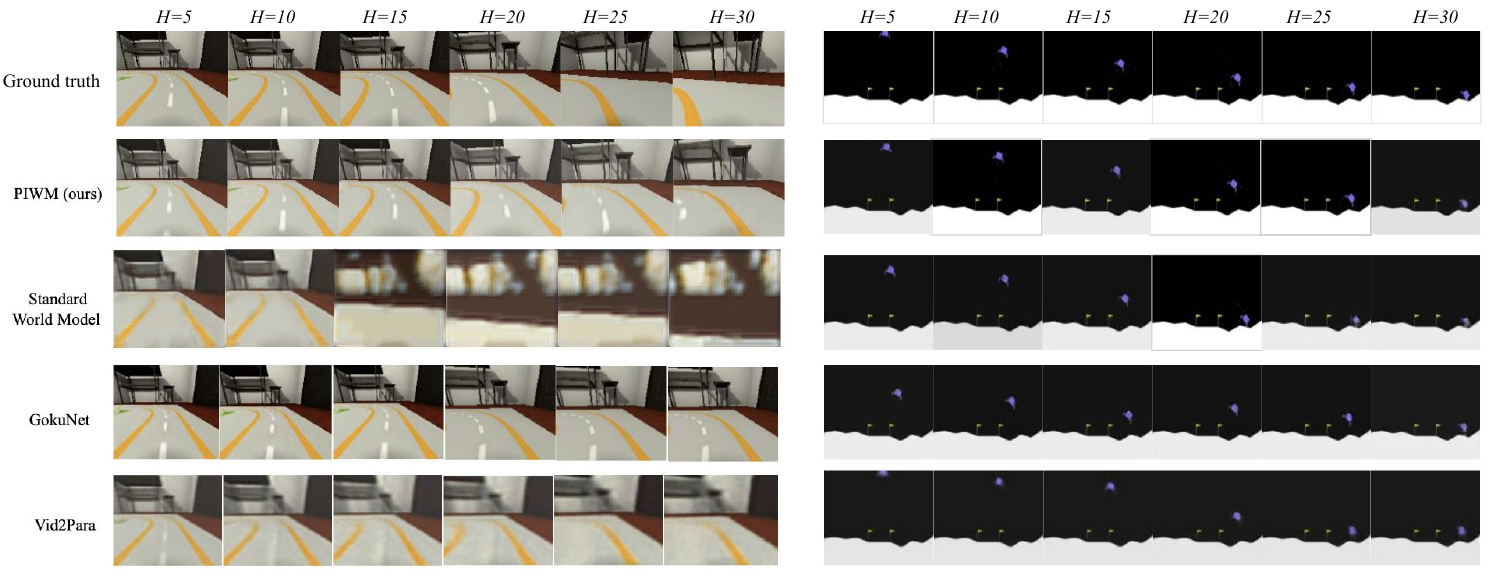}
\caption{Qualitative comparison of 30-step trajectory rollouts under $\delta=5\%$.}	\label{fig:4}

 \end{figure*}

\paragraph{Predictive performance and interpretability.}
Our experiments show that the extrinsic architecture with a discrete latent space is optimal for learning physically interpretable world models from weak supervision. This approach achieves superior prediction accuracy (Figure~\ref{fig:2}) by decoupling perception from physical state abstraction and leveraging quantization as a powerful regularizer against visual noise. The learned representations are not only predictive but also genuinely physically grounded, as evidenced by the model's ability to recover true system parameters (Figure~\ref{fig:3}) and generate qualitatively plausible visual rollouts that far exceed baseline performance (Figure~\ref{fig:4}). Crucially, the substantial gains in interpretability and prediction accuracy come at a minimal cost to downstream controller performance, as reported in Table~\ref{tab:2} in the Appendix. Across all variants and noise levels, predicted actions are largely unperturbed, confirming that physical interpretability does not degrade task-relevant visual fidelity. Notably, the extrinsic-discrete variant achieves the strongest balance between physical grounding and reconstruction quality for downstream control. DVBF performs poorly in our settings because its Gaussian latent space and amortized inference provide no physical inductive bias, causing the learned states to drift and quickly accumulate errors during multi-step rollouts. This makes DVBF unstable in visually rich or discontinuous dynamics environments, which explains its large performance gap relative to structured models such as PIWM. We note that all baselines, including the standard world model, were trained to convergence using early stopping on the same validation set, ruling out underfitting as an explanation for the performance gap.

\paragraph{Out-of-distribution generalization.}
It remains an open question whether the learned physically interpretable representations exhibit domain transfer and better compositional generalization than purely data-driven baselines, for example, under unseen initial conditions, modified system parameters, or visual domain shifts. We expect physical grounding to provide a structural advantage in such scenarios, but systematic evaluation is left for future work. Since PIWM currently requires a partially specified dynamics model, it would be promising to discover the equation structure, for example, via symbolic regression or neural ODEs.

Our framework assumes that the partially known dynamics model provides a reasonable approximation of the true system. When the assumed equations omit significant physical effects (e.g., tire slip, air resistance, or actuator delays), the learned parameters may compensate to some extent, but prediction accuracy will degrade. Investigating graceful degradation under deliberately misspecified dynamics, and developing mechanisms to detect or correct such misspecification, is an important direction for future work.

\paragraph{Long-term dependencies.} 
\looseness=-1
Our experiments focused on Markovian dynamical systems, where transitions depend only on the most recent latent state and action. This setting aligns with the majority of prior work on world models and enables a clean evaluation. Extending PIWM to non-Markovian or delayed-effect systems, where temporal dependencies span multiple steps, is an interesting direction for future research. Evaluating such scenarios would require augmenting the dynamics model with additional memory mechanisms or explicitly modeling higher-order temporal dependencies. Furthermore, the rich temporal nature of weak supervision signals is currently underutilized. Future methods could process sequences of noisy supervisory signals using filtering or sequence modeling techniques to produce a more refined, temporally coherent learning target, thereby improving the model's robustness and accuracy. 

\paragraph{Open-world systems.}
\looseness=-1
While CartPole and Lunar Lander provide controlled benchmarks, DonkeyCar represents a real robotic platform with visual sensing, approximate physical modeling, and closed-loop control. Scaling PIWM to open-world CPS scenarios, such as multi-agent driving environments with richer state spaces, constitutes an important direction for future work. While our framework is effective, future work should also focus on scaling its representational capacity and enhancing its use of supervision. For complex, open-world scenarios like autonomous driving, our approach could be extended from predicting simple state vectors to building structured world representations, such as dynamic 3D occupancy grids, where physical priors can be applied to multiple agents.

\section{Related Work}\label{related}

\subsection{Trajectory Prediction}


Predicting trajectories is critical for safe planning and control~\cite[]{fridovich2020confidence}, but existing methods present trade-offs. While approaches like Hamilton-Jacobi (HJ) reachability offer formal guarantees~\cite[]{li2021prediction,nakamura2023online}, they are computationally expensive for online settings. Conversely, mainstream deep learning models are powerful but often rely on handcrafted scene representations~\cite[]{salzmann2020trajectron++} or high-precision maps~\cite[]{itkina2023interpretable, hsu2023interpretable}, and typically do not produce physically interpretable predictions~\cite[]{lu2024towards, lindemann2023conformal, ruchkin2022confidence}. In contrast, our approach learns directly from raw images with distribution-based weak supervision without requiring handcrafted inputs or goal conditioning.

\subsection{Representation Learning}


A central challenge in learning from high-dimensional sequences is creating compact and meaningful latent representations~\cite[]{shi2015convolutional,bai2018empirical}. While Variational Autoencoders (VAEs)~\cite[]{kingma2013auto} are foundational, their standard form learns unstructured latents. A significant line of research attempts to impose structure, either by encouraging disentanglement in continuous spaces with methods like \(\beta\)-VAE, FactorVAE, and TCVAE~\cite[]{higgins2017beta, kim2018disentangling, chen2018isolating}, or by enforcing causality~\cite[]{yang2021causalvae}. However, these methods often fall short of ensuring a direct correspondence with real-world physical states or require precisely labeled data. An alternative is to impose structure via discretization with Vector-Quantized VAEs (VQ-VAEs) and their extensions~\cite[]{van2017neural, razavi2019generating, xue2019supervised}. While effective, their application to physical prediction has been limited due to abstract latent codes and reliance on exact supervision. Even in robotic applications like DVQ-VAE that model structured systems, encoding external environmental factors remains a challenge~\cite[]{zhao2024decomposed}. GOKU-net constrains variables to plausible physical ranges but does not tie them to specific, interpretable quantities~\cite[]{linial_generative_2021}.

Other works incorporate structured priors to enhance learning. Object-centric world models like FOCUS~\cite[]{ferraro2023focus} and foundation world models~\cite{mao2024zero} improve efficiency by structuring latents around discrete entities. Models like SPARTAN~\cite []{lei2024spartan} provide outputs that are interpretable by construction but do not incorporate known physical dynamics or action-conditioned prediction. Priors can also be introduced as task-related rules in Bayesian neural networks~\cite[]{sam2024bayesian} or through human and self-supervision to guide abstraction~\cite[]{fu2021learning, wen2023any, konidaris2018skills, peng2024human, chen2022automated}. These approaches, however, often yield abstract representations that lack physical grounding without strong external priors. Closer to our work, Deep Variational Bayes Filters (DVBFs) extend VAEs with a latent dynamics module~\cite[]{karl2016deep}. Yet, without explicit physical supervision, they often fail to recover interpretable variables --- a limitation our work addresses by leveraging weak supervision.

\subsection{World Models}


Classical world models like Dreamer~\cite[]{hafner2020mastering} and DayDreamer~\cite[]{wu2023daydreamer} excel at learning from experience for policy learning, but their latent representations are typically uninterpretable~\cite[]{peper2025four}. Many approaches seek to improve physical grounding by incorporating priors, such as bounds on states and actions~\cite[]{tumu2023physics,sridhar2023guaranteed}, physics-aware loss functions~\cite[]{djeumou2023learn}, or kinematics-inspired layers~\cite[]{cui2020deep}. However, these methods are often designed for low-dimensional systems and do not scale well to learning from noisy, high-dimensional images. Other physics-informed methods leverage differential equations to stabilize learning~\cite[]{ZHONG2023115664,linial_generative_2021}, with frameworks like Phy-Taylor using Taylor monomials to structure the latent dynamics~\cite[]{mao_phy-taylor_2025}. Approaches like sparse identification and differentiable physics require access to the underlying state variables and are not designed to learn from raw visual inputs~\cite[]{yao2024marrying,BRUNTON2016710,de2018end}.

Recent advances have produced powerful but distinct world models. For instance, 3D occupancy-based models improve forecasting but their internal states are not explicitly aligned with physical variables~\cite[]{zheng2024occworld,min2023uniworld,yan2024renderworld,zuo2024gaussianworld}. Concurrently, neuro-symbolic models enhance generalization but require predefined symbolic inputs not available from raw sensor data~\cite[]{balloch2023neuro,liang2024visualpredicator}. Our approach is distinct from these paradigms as it learns physically interpretable representations directly from images using weak supervision. This also contrasts with the most closely related work, Vid2Param~\cite[]{asenov2019vid2param}, which requires full supervision and struggles with dynamics prediction.


\section{Conclusion}\label{sec:conclusion}
We presented the Physically Interpretable World Model, a framework that learns physically-grounded latent representations from images using only weak distributional supervision. Our systematic evaluation demonstrates that an extrinsic architecture with a discrete latent space yields accurate and robust predictions and successfully recovers the system's true physical parameters. This work not only provides direct evidence of a physically interpretable model --- but also offers a practical path toward more trustworthy and reliable cyber-physical systems with generative predictors.

\section*{Acknowledgements}\label{sec:ack}
 
The authors thank Liam (Cade) McGlothlin and Vedansh Maheshwari for their help in exploring interpretable world models. The authors also appreciate the feedback of the anonymous reviewers. 

This material is based upon work supported by the U.S. National Science Foundation (NSF) under Grant Numbers CNS 2513076 and CCF 2403616. Any opinions, findings, and conclusions or recommendations expressed in this material are those of the authors and do not necessarily reflect the views of the NSF.


\bibliographystyle{ACM-Reference-Format}
\bibliography{sample-base}

\newpage

\clearpage

\appendix
\section*{Appendix}

\textbf{Model Architecture and Training Details.} This appendix provides detailed implementation specifications for the Physically
Interpretable World Model (PIWM) and its four instantiations:
intrinsic-continuous, intrinsic-discrete, extrinsic-continuous, and
extrinsic-discrete. All variants follow the unified formulation introduced in
Section~\ref{sec:architecture}, operating on the same dataset, the same weak
supervision mechanism, and the same training objective.

\subsubsection*{A. Vision Encoders and Decoders}

All variants share the same convolutional vision encoder $\mathcal{E}_v$ and
decoder $\mathcal{D}_v$. The encoder contains two convolutional layers with ReLU
activations and maps each observation $y_t$ to an intermediate latent vector
$z_t \in \mathbb{R}^{64}$. The decoder $\mathcal{D}_v$ mirrors this structure
and reconstructs an image $\hat{y}_{t+2}$ from a latent representation.  
Continuous variants use a standard $\beta$-VAE objective on $z_t$ with
$\beta = 1$, while discrete variants apply a VQ-VAE quantization layer with a
codebook of 512 learnable embeddings of dimension 64, trained using a
commitment weight of $\beta = 0.25$.

\subsubsection*{B. Physical Encoder $\mathcal{E}_p$}

In extrinsic architectures, the interpretable latent state is produced by a
physical encoder $\mathcal{E}_p$ applied to the intermediate visual latent:
\[
z^*_t = \mathcal{E}_p(z_t).
\]
This module is implemented as a Transformer encoder with two layers, four
attention heads per layer, a model dimension of 128, and a feedforward width of
512. Mean pooling across tokens is followed by a linear projection to produce
the final interpretable state $z^*_t$. The physical encoder is trained entirely
via the interpretability loss $\mathcal{L}_{\text{interp}}(z^*_t, \Xi_t)$.

\subsubsection*{C. Intrinsic Representation Learning}

Intrinsic variants produce the interpretable latent state directly from images,
\[
z^*_t = \mathcal{E}(y_t),
\]
without an intermediate latent or a separate physical encoder.  
For intrinsic--continuous models, $\mathcal{E}$ outputs the parameters of a
Gaussian distribution over the latent $z^*_t \in \mathbb{R}^{64}$, where a fixed
subset of dimensions corresponds to the physical component $z^{*}_{p,t}$.  
The loss combines reconstruction, interpretability alignment, and KL
regularization following the $\beta$-VAE formulation.

For intrinsic--discrete models, a VQ-VAE codebook of 512 embeddings is used.
Each embedding vector is partitioned as
$\mathbf{e}_k = [\mathbf{e}^{p}_k,\, \mathbf{e}^{v}_k]$, matching the structure
in Section~\ref{sec:architecture}. The interpretable state is computed from the
physical portion of the selected codebook entry, and the interpretability loss is
applied directly to this physical component.

\subsubsection*{D. Dynamics Model}

All variants share the same structured second-order latent dynamics model
described in Section~\ref{sec:architecture}:
\[
z^{*}_{t+2} = \phi_\theta(z^{*}_{t},\, z^{*}_{t+1},\, a_t).
\]
The functional form of $\phi_\theta$ is derived from the physics of each
environment (e.g., kinematics), with only the parameters $\theta$ being
learnable. Depending on the environment, $\phi_\theta$ contains between 4 and 10
unknown physical parameters. A two-step initialization window
$(z^*_t, z^*_{t+1})$ provides the required derivative information for
prediction, such as implicit velocities.  

At each time step, weak supervision consists of a set of $L = 50$ proxy samples
$\Xi_t = \{\xi^{(l)}_t\}$ drawn from the unknown distribution $p(x_t)$. The
dynamics loss is computed as
\[
\mathcal{L}_{\text{dyn}} = 
\| z^{*}_{t+H} - \hat{\mu}_{\xi_{t+H}} \|_2^2,
\]
where $\hat{\mu}_{\xi_t}$ is the empirical sample mean of $\Xi_t$.

\subsubsection*{E. Loss Functions}

All variants are trained using the unified objective:
\[
\mathcal{L} =
\mathcal{L}_{\text{rec}}
+ \lambda_1 \mathcal{L}_{\text{interp}}
+ \lambda_2 \mathcal{L}_{\text{dyn}}
+ \lambda_3 \mathcal{L}_{\text{reg}}.
\]
The reconstruction loss is mean squared error in image space:
\[
\mathcal{L}_{\text{rec}}
= \| \mathcal{D}_v(z_{t+2}) - y_{t+2} \|_2^2.
\]
The interpretability loss uses the weak supervision samples:
\[
\mathcal{L}_{\text{interp}}
= \| z^{*}_{p,t} - \hat{\mu}_{\xi_t} \|_2^2.
\]

Continuous variants use KL regularization consistent with the $\beta$-VAE
formulation, whereas discrete variants use VQ-VAE codebook and commitment
losses, matching the structure defined in Section~\ref{sec:architecture}.

\subsubsection*{F. Training Setup}

All models are trained using the Adam optimizer with batch size 32. Continuous
variants use an initial learning rate of $10^{-4}$, and discrete variants use
$10^{-3}$ due to the stochasticity introduced by quantization. A cosine decay
schedule with 5 warmup epochs is applied. Training proceeds for up to 200
epochs, with early stopping after 20 epochs of no validation improvement.  
Gradient norms are clipped at 1.0 for stability.  
Evaluation uses 30-step prediction rollouts in latent space, decoding to images
only when needed.

Weak supervision is applied consistently across noise settings
$\delta \in \{0\%, 5\%, 10\%\}$.

\subsubsection*{G. Summary of Differences Between Variants}

Intrinsic models produce $z^{*}_t$ directly from observations, whereas extrinsic
models obtain $z^{*}_t$ through the composition
$z^*_t = \mathcal{E}_p(\mathcal{E}_v(y_t))$.  
Continuous variants use Gaussian latent parameterizations, and discrete variants
use a VQ-VAE codebook with 512 entries.  
Across all variants, the interpretable component of the latent occupies a fixed
subset of dimensions. All models share identical dynamics functions
$\phi_\theta$ and training losses.

\subsubsection*{H. Architectural Details in Text}

All models use a shared two-layer CNN vision encoder and decoder producing a
64-dimensional latent space. Continuous variants use Gaussian latents with a KL
penalty, while discrete variants use a 512-entry VQ-VAE codebook trained with a
commitment weight of $\beta=0.25$.  
Extrinsic variants include a two-layer Transformer physical encoder with model
dimension 128. Intrinsic variants instead embed the physical portion directly
into $z^*_t$.  
The dynamics model learns between 4 and 10 environment-specific parameters using
50 weak supervision samples per time step.

\subsubsection*{I. Training Hyperparameters in Text}

All experiments use batch size 32, cosine learning-rate decay with 5 warmup
epochs, a maximum of 200 epochs with patience 20, and gradient clipping at
1.0. Rollouts span 30 steps in latent space. All environments use the same
60{,}000-trajectory dataset and weak supervision noise levels
$\delta \in \{0\%, 5\%, 10\%\}$.

\begin{figure*}[!htbp]
    \centering
    \includegraphics[width=1.95\columnwidth]{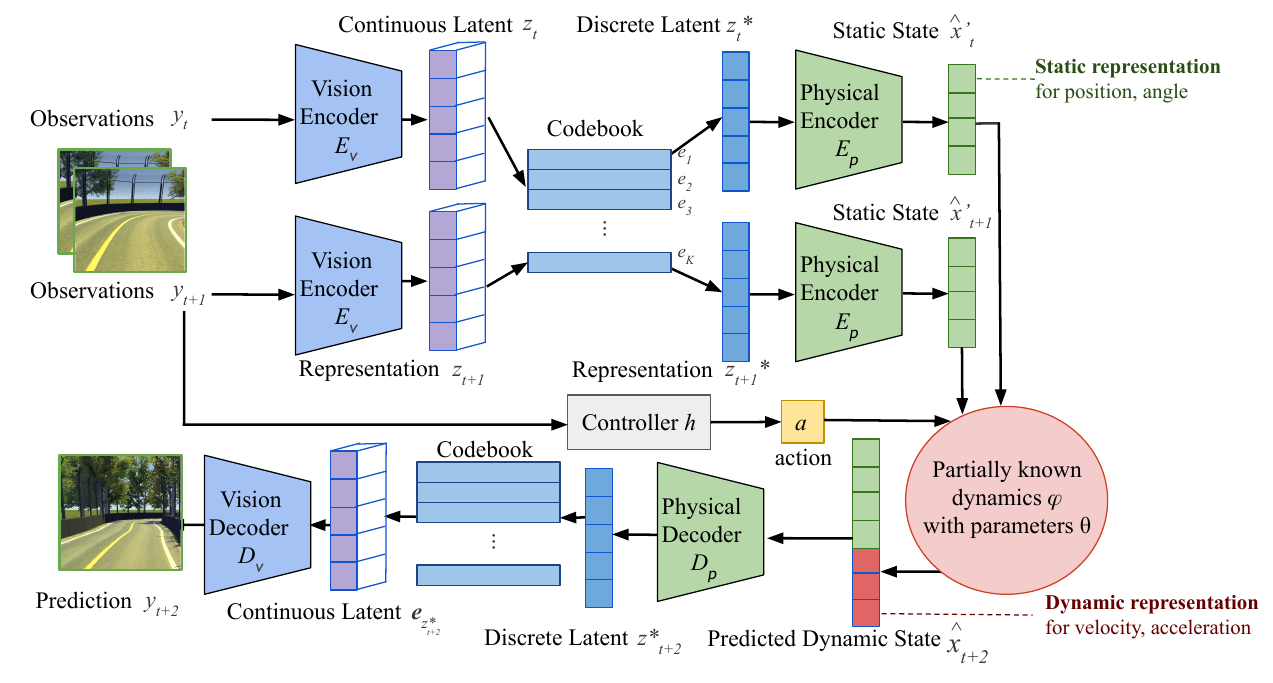}
\caption{
\textbf{Extrinsic–discrete PIWM architecture.}
The vision encoder $\mathcal{E}_v$ produces latents $z_t$, which are quantized
to $z_t^{*}$ and mapped to physical states by $\mathcal{E}_p$. The dynamics
model $\phi_\theta$ predicts $\hat{x}_{t+2}$ from two physical states and
action $a_t$, and the decoders $\mathcal{D}_p$ and $\mathcal{D}_v$ reconstruct
the predicted observation.
}
    \label{fig:extrinsic-vq-appendix}
\end{figure*}

\begin{figure}[h]
	\centering         \includegraphics[width=0.95\columnwidth]{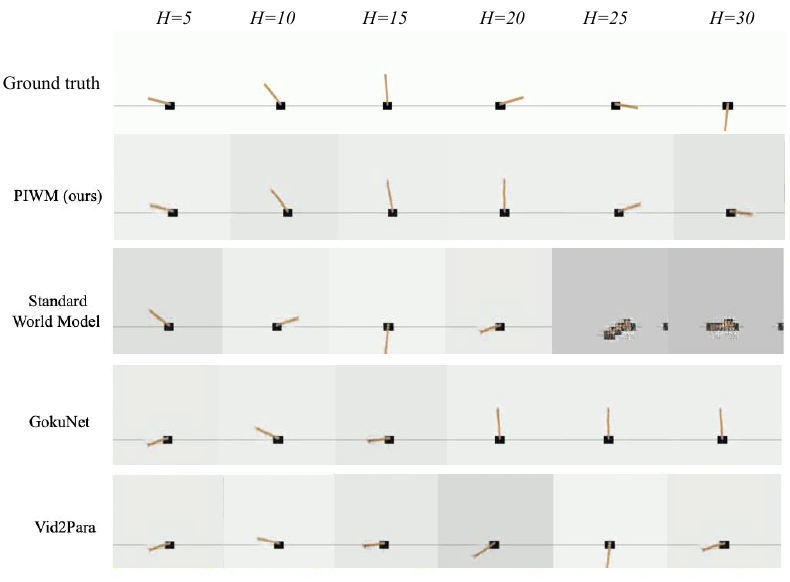}
\caption{Qualitative comparison of 30-step trajectory rollouts under $\delta=5\%$ of Cart Pole.}	\label{fig:cartviz}
 \end{figure}

\textbf{Architecture for the Extrinsic-Discrete Variant}. 
This section provides an expanded diagram of the \emph{extrinsic-discrete}
variant of PIWM, which is used in our experiments.  
Unlike intrinsic variants, the extrinsic pipeline decouples visual perception
from physical interpretation: a VQ-based vision autoencoder first produces a
discrete latent representation, after which a dedicated physical encoder
extracts the interpretable physical state.  
This figure visualizes the complete data flow from raw observations to discrete
latents, physical state estimation, structured dynamics prediction, and
final reconstruction.

\textbf{Controller Performance on Reconstructions.}
To assess whether PIWM variants preserve task-relevant visual information, we
evaluate how well reconstructed observations $\hat{y}_{t}$ support action
prediction when passed through the environment’s original controller $h$.  
Table~\ref{tab:2} summarizes controller performance (action RMSE
for Donkey Car; action accuracy for Lunar Lander and CartPole) for all four
architectural variants and all supervision noise levels.  
These results provide a complementary diagnostic showing how the different
latent designs affect the quality of reconstructed observations for downstream
control.

Algorithms~\ref{alg:pendulum_dynamics},~\ref{alg:lunar_lander_dynamics}, and~\ref{alg:bicycle_dynamics} describe the dynamics equations used as the partially known dynamics model $\phi_\theta$ in each environment.

Algorithm~\ref{alg:pendulum_dynamics} models the Cart Pole system, where the state consists of cart position $x$, cart velocity $\dot{x}$, pole angle $\theta$, and angular velocity $\dot{\theta}$. The learnable parameters include the pole mass $m_p$, pole length $l$, cart mass $m_c$, and applied force magnitude.

Algorithm~\ref{alg:lunar_lander_dynamics} presents the Lunar Lander dynamics with a six-dimensional state: position $(x, y)$, velocity $(\dot{x}, \dot{y})$, orientation $\theta$, and angular velocity $\dot{\theta}$. The learnable parameters are the main engine power and side engine power.

Algorithm~\ref{alg:bicycle_dynamics} defines a simplified bicycle model used to approximate the Donkey Car dynamics. The state consists of position $(x, y)$, heading $\theta$, and speed $v$. The primary learnable parameter is the wheelbase $L$. Since the true Donkey Car physics are governed by the Unity simulator, ground-truth parameters are unavailable for this environment.

\begin{table}[ht!]
\centering
\caption{Controller Performance on Reconstructed Observations Across All Variants and Noise Levels}
\label{tab:2}
\tiny 
\begin{tabular}{@{}ll l ccc@{}}
\toprule
\multirow{2}{*}{\textbf{Case}} & \multirow{2}{*}{\textbf{Input Type}} & \multirow{2}{*}{\textbf{Latent Space}} & \multicolumn{3}{c}{\textbf{Supervision Noise Level ($\delta$)}} \\
\cmidrule(l){4-6}
& & & \textbf{0\%} & \textbf{5\%} & \textbf{10\%} \\
\midrule
\multirow{5}{*}{\shortstack{Donkey Car \\ (Action RMSE $\downarrow$)}}
& \multirow{2}{*}{One-Stage (Intrinsic)} & Continuous & $0.12 \pm 0.04$ & $0.13 \pm 0.04$ & $0.15 \pm 0.05$ \\
& & Discrete & $0.21 \pm 0.15$ & $0.29 \pm 0.16$ & $0.32 \pm 0.20$ \\
\cmidrule(l){2-6}
& \multirow{2}{*}{Two-Stage (Extrinsic)} & Continuous & $0.15 \pm 0.05$ & $0.16 \pm 0.05$ & $0.22 \pm 0.06$ \\
& & Discrete & $0.15 \pm 0.04$ & $0.17 \pm 0.05$ & $0.19 \pm 0.05$ \\
\midrule
\multirow{5}{*}{\shortstack{Lunar Lander \\ (Action Acc. $\uparrow$)}}
& \multirow{2}{*}{One-Stage (Intrinsic)} & Continuous & $93.0\% \pm 1.8\%$ & $90.5\% \pm 2.0\%$ & $87.1\% \pm 2.2\%$ \\
& & Discrete & $85.5\% \pm 2.5\%$ & $82.1\% \pm 2.8\%$ & $78.3\% \pm 3.1\%$ \\
\cmidrule(l){2-6}
& \multirow{2}{*}{Two-Stage (Extrinsic)} & Continuous & $86.2\% \pm 2.4\%$ & $83.5\% \pm 2.6\%$ & $80.0\% \pm 2.9\%$ \\
& & Discrete & $91.5\% \pm 2.1\%$ & $88.6\% \pm 2.3\%$ & $84.5\% \pm 2.5\%$ \\
\midrule
\multirow{5}{*}{\shortstack{Cart Pole \\ (Action Acc. $\uparrow$)}}
& \multirow{2}{*}{One-Stage (Intrinsic)} & Continuous & $98.0\% \pm 1.0\%$ & $96.5\% \pm 1.2\%$ & $94.0\% \pm 1.5\%$ \\
& & Discrete & $95.0\% \pm 1.6\%$ & $91.5\% \pm 2.0\%$ & $87.2\% \pm 2.5\%$ \\
\cmidrule(l){2-6}
& \multirow{2}{*}{Two-Stage (Extrinsic)} & Continuous & $95.5\% \pm 1.5\%$ & $92.0\% \pm 1.8\%$ & $88.0\% \pm 2.2\%$ \\
& & Discrete & $97.2\% \pm 1.1\%$ & $95.0\% \pm 1.4\%$ & $92.5\% \pm 1.8\%$ \\
\bottomrule
\end{tabular}
\end{table}


\newpage

\begin{algorithm}[h]
\caption{Dynamics of Cart Pole}
\label{alg:pendulum_dynamics}
\begin{algorithmic}[1]
\Require Current state $z = [x, \dot{x}, \theta, \dot{\theta}]$, action $a$
\Ensure  Updated state $z_{\text{new}} = [x_{\text{new}}, \dot{x}_{\text{new}}, \theta_{\text{new}}, \dot{\theta}_{\text{new}}]$

\State \textbf{// Extract State Variables}
\State $x,\ \dot{x},\ \theta,\ \dot{\theta}
       \gets z[:,0],\ z[:,1],\ z[:,2],\ z[:,3]$

\State \textbf{// Convert Action to Force}
\State $F \gets \text{force\_mag} \times (2 \cdot a - 1)$

\State \textbf{// Compute Trigonometric Values}
\State $\cos\theta \gets \text{costheta},\quad \sin\theta \gets \text{sintheta}$

\State \textbf{// Compute Intermediate Variable}
\State $\text{temp} \gets
       \dfrac{F + m_p \cdot l \cdot \dot{\theta}^2 \cdot \sin\theta}
             {m_p + m_c}$

\State \textbf{// Calculate Angular Acceleration}
\State $\ddot{\theta} \gets
       \dfrac{g \cdot \sin\theta - \cos\theta \cdot \text{temp}}
             {l \cdot \left(\dfrac{4}{3}
              - \dfrac{m_p \cdot \cos^2\!\theta}{m_p + m_c}\right)}$

\State \textbf{// Calculate Linear Acceleration}
\State $\ddot{x} \gets \text{temp}
       - \dfrac{m_p \cdot l \cdot \ddot{\theta} \cdot \cos\theta}{m_p + m_c}$

\State \textbf{// Update State Variables}
\State $x_{\text{new}}         \gets x         + \tau \cdot \dot{x}$
\State $\dot{x}_{\text{new}}   \gets \dot{x}   + \tau \cdot \ddot{x}$
\State $\theta_{\text{new}}    \gets \theta     + \tau \cdot \dot{\theta}$
\State $\dot{\theta}_{\text{new}} \gets \dot{\theta} + \tau \cdot \ddot{\theta}$

\State \Return
       $z_{\text{new}} \gets
       [x_{\text{new}},\ \dot{x}_{\text{new}},\
        \theta_{\text{new}},\ \dot{\theta}_{\text{new}}]$
\end{algorithmic}
\end{algorithm}


\newpage

\begin{algorithm}[h]
\caption{Dynamics of Lunar Lander}
\label{alg:lunar_lander_dynamics}
\begin{algorithmic}[1]
\Require Current states
         $\mathbf{s} = [x, y, \dot{x}, \dot{y}, \theta, \dot{\theta}]$,
         actions $\mathbf{a}$
         \textit{(0: do nothing, 1: fire left, 2: fire main, 3: fire right)}
\Ensure  Updated states
         $\mathbf{s}_{\text{new}} =
         [x_{\text{new}}, y_{\text{new}},
          \dot{x}_{\text{new}}, \dot{y}_{\text{new}},
          \theta_{\text{new}}, \dot{\theta}_{\text{new}}]$

\State \textbf{// Unpack State Variables}
\State $x \gets \mathbf{s}[:,0],\quad y \gets \mathbf{s}[:,1]$
\State $\dot{x} \gets \mathbf{s}[:,2],\quad \dot{y} \gets \mathbf{s}[:,3]$
\State $\theta \gets \mathbf{s}[:,4],\quad \dot{\theta} \gets \mathbf{s}[:,5]$

\State \textbf{// Calculate Engine Direction and Dispersion}
\State $\text{tip}  \gets [\sin\theta,\ \cos\theta]$
\State $\text{side} \gets [-\cos\theta,\ \sin\theta]$

\State \textbf{// Process Actions}
\State $\text{fire\_main}  \gets (\mathbf{a} == 2)$
\State $\text{fire\_left}  \gets (\mathbf{a} == 1)$
\State $\text{fire\_right} \gets (\mathbf{a} == 3)$

\State \textbf{// Compute Main Engine Thrust}
\State $m_{\text{power}} \gets \text{fire\_main}$
\State $\dot{x} \gets \dot{x}
       - \text{tip}[:,0] \cdot \text{main\_power}
         \cdot m_{\text{power}} \;/\; \text{FPS}$
\State $\dot{y} \gets \dot{y}
       + \text{tip}[:,1] \cdot \text{main\_power}
         \cdot m_{\text{power}} \;/\; \text{FPS}$

\State \textbf{// Compute Side Engine Thrust}
\State $s_{\text{power}}  \gets \text{fire\_left} + \text{fire\_right}$
\State $\text{direction}  \gets \text{fire\_right} - \text{fire\_left}$
\State $\dot{x} \gets \dot{x}
       + \text{side}[:,0] \cdot \text{side\_power}
         \cdot s_{\text{power}} \cdot \text{direction} \;/\; \text{FPS}$
\State $\dot{\theta} \gets \dot{\theta}
       + \text{side\_power} \cdot s_{\text{power}}
         \cdot \text{direction} \;/\; \text{FPS}$

\State \textbf{// Update Position and Angle}
\State $x      \gets x      + \dot{x}      / \text{FPS}$
\State $y      \gets y      + \dot{y}      / \text{FPS}$
\State $\theta \gets \theta + \dot{\theta} / \text{FPS}$

\State \Return
       $\mathbf{s}_{\text{new}} \gets
       [x,\ y,\ \dot{x},\ \dot{y},\ \theta,\ \dot{\theta}]$
\end{algorithmic}
\end{algorithm}

\begin{algorithm}[h]
\caption{Dynamics of Bicycle Model}
\label{alg:bicycle_dynamics}
\begin{algorithmic}[1]
\Require Current state $s = [x, y, \theta, v]$,
         action $a = [\delta,\, a_{\text{acc}}]$
\Ensure  Updated state
         $s_{\text{new}} =
         [x_{\text{new}}, y_{\text{new}}, \theta_{\text{new}}, v_{\text{new}}]$

\State \textbf{// Extract State Variables}
\State $x,\ y,\ \theta,\ v
       \gets s[:,0],\ s[:,1],\ s[:,2],\ s[:,3]$

\State \textbf{// Extract Action Variables}
\State $\delta,\ a_{\text{acc}} \gets a[:,0],\ a[:,1]$

\State \textbf{// Update State Variables}
\State $x_{\text{new}}      \gets x      + v \cdot \cos(\theta) \cdot \tau$
\State $y_{\text{new}}      \gets y      + v \cdot \sin(\theta) \cdot \tau$
\State $\theta_{\text{new}} \gets \theta + \dfrac{v}{L} \cdot \tan(\delta) \cdot \tau$
\State $v_{\text{new}}      \gets v      + a_{\text{acc}} \cdot \tau$

\State \Return
       $s_{\text{new}} \gets
       [x_{\text{new}},\ y_{\text{new}},\
        \theta_{\text{new}},\ v_{\text{new}}]$
\end{algorithmic}
\end{algorithm}

\end{document}